\documentclass[preprint,12pt]{elsarticle}

\usepackage{amssymb}

\usepackage{subcaption}
\usepackage{booktabs}
\usepackage{multirow}
\usepackage{pifont} 
\usepackage{subfig}
\usepackage{float}

\usepackage{subcaption}
\usepackage{booktabs}
\usepackage{multirow}
\usepackage{pifont} 
\usepackage{subfig}
\usepackage{soul}
\usepackage{xcolor}
\definecolor{myhighlight}{RGB}{255,215,0}
\usepackage{float}
\usepackage{booktabs} 
\usepackage{array} 
\usepackage{multirow} 

\journal{arXiv.org}

\begin{document}

\begin{frontmatter}



\title{Biasing \& Debiasing based Approach Towards Fair Knowledge Transfer for Equitable Skin Analysis}


\author{Anshul Pundhir\corref{corauthor}}             
		\ead{anshul\_p@cs.iitr.ac.in} 
  \author{Balasubramanian Raman}
  \author{Pravendra Singh}
             
            \address{\textnormal{\{anshul\_p, bala, pravendra.singh\}@cs.iitr.ac.in}\\ \vspace{0.2cm} Department of Computer Science and Engineering\\Indian Institute of Technology Roorkee, Roorkee, Uttarakhand, India - 207001
            }
            
            

  \cortext[corauthor]{Corresponding author}
  
\begin{abstract}
Deep learning models, particularly Convolutional Neural Networks (CNNs), have demonstrated exceptional performance in diagnosing skin diseases, often outperforming dermatologists. However, they have also unveiled biases linked to specific demographic traits, notably concerning diverse skin tones or gender, prompting concerns regarding fairness and limiting their widespread deployment. Researchers are actively working to ensure fairness in AI-based solutions, but existing methods incur an accuracy loss when striving for fairness. To solve this issue, we propose a `two-biased teachers' (i.e., biased on different sensitive attributes) based approach to transfer fair knowledge into the student network. Our approach mitigates biases present in the student network without harming its predictive accuracy. In fact, in most cases, our approach improves the accuracy of the baseline model. To achieve this goal, we developed a weighted loss function comprising biasing and debiasing loss terms. We surpassed available state-of-the-art approaches to attain fairness and also improved the accuracy at the same time. The proposed approach has been evaluated and validated on two dermatology datasets using standard accuracy and fairness evaluation measures. We will make source code publicly available to foster reproducibility and future research.   
\end{abstract}




\end{frontmatter}


\section{Introduction and Related Works} 
With recent advancements in healthcare, computer-aided diagnosis (CAD) has become a common choice by medical practitioners as it reduces the manual burden, improves accuracy, and is also helpful in assisting the less experienced doctors to make their decision~\cite{chen2022representative,goceri2023classification,zhou2023fixmatch}. Due to the excellent data distribution learning tendency in Deep Neural Networks (DNN), they are a common choice while building CAD systems. To provide excellent performance, DNN also learns sensitive attributes like colors, gender, skin tone, and textures in the training data, which act as  bias~\cite{du2021fairness,lu2021evaluating,li2022bias,liu2023hybrid}. Due to learning of unwanted features or biased information, DNN-based systems show biased performance during inference across the population subgroups. For example, researchers have shown that DNN leads to biased performance across demographic groups while predicting attributes~\cite{du2021fairness,xu2020investigating, petersen2022feature,mangotra2023effect}. For X-ray-based CAD, researchers have highlighted the issue of gender bias, which severely affects the model performance and leads to injustice for sub-population groups~\cite{larrazabal2020gender}.\\Recently, authors have highlighted the issue of skin tone bias in dermatology, which leads to unfair diagnosis systems in dermatology~\cite{groh2021evaluating,kinyanjui2020fairness}. Due to biased DNN models, it is challenging to wide-spread their deployment for real-world applications. The pressing need for fair AI models to solve the bias issue to overcome unjust discrimination across different subgroups motivates us to come up with fair and efficient solutions to solve these challenges. Researchers widely employed adversarial approaches to mitigate bias during model training~\cite{kim2019learning, wang2019balanced, elazar2018adversarial, alvi2018turning, zhang2018mitigating}. To solve the bias issue, the adversarial training approaches follow the min-max optimization objective, where the network aims to minimize the loss in the primary objective of target classification but maximize the auxiliary task of bias prediction.
Researchers found that the adversarial training-based approaches are still not robust enough to provide fair representations as they harm the classification rate on target tasks, and still, some feature combinations contain the sensitive feature influence~\cite{wang2020towards}.\\ 
\noindent
To solve data bias, researchers generally use data-processing techniques, which include pre-processing, in-processing, and post-processing. Pre-processing techniques are used to make the dataset learning fair by suppressing the influence of biased information by weighting the training samples~\cite{lu2020gender, ngxande2020bias}. The in-processing techniques are employed to provide fairness by making changes in network architecture and loss functions and following certain training strategies so as to mitigate bias attribute learning during model training~\cite{jung2021fair, quadrianto2019discovering}. The post-processing techniques make calibration by making use of sensitive information and model prediction so as to enhance fairness by aligning the distribution learned by the model~\cite{du2020fairness, hardt2016equality, zhao2017men}.\\
Recently, researchers have been paying attention to solving the data bias issue using several methods. For example, to solve the bias issue in the text-to-image diffusion model, authors have used distribution alignment loss~\cite{shen2023finetuning}. Qraitem et al.~\cite{qraitem2023bias} proposed a bias-mimicking sampling approach to mitigate bias issues. Dhar et al.~\cite{dhar2021distill} proposed a knowledge distillation-based approach to mitigate bias during face verification where the student is forced to learn unbiased representation from a teacher using feature-level distillation loss, and student teachers are given samples of different sensitive attributes. To ensure fairness during knowledge transfer from teacher to student network, Yue et al.~\cite{yue2024revisiting} proposed adversarial robust distillation by increasing the weights of difficult classes in student knowledge transfer. In a similar realm, Dong et al. ~\cite{dong2023reliant} ensure fairness during knowledge transfer in graph neural networks. Li et al.~\cite{li2023dual} solve the multi-domain bias issue in fake news detection by knowledge distillation in the form of adversarial debiasing distillation loss from the unbiased teacher so that student model learn unbiased domain knowledge.\\For dermatology datasets, researchers are actively solving issues due to skin tone bias and gender bias. Xu et al.~\cite{xu2023fairadabn} attempted to solve the fairness issue and consider the statistical distribution of subgroups present in the dataset and proposed modifying the model architecture by replacing the batch normalization layer with their proposed normalization layer (FairAdaBN). Chiu et al.~\cite{chiu2023toward} developed a multi-exit framework that solves the fairness issue using shallow layer features. Recently, Wu et al.~\cite{wu2022fairprune} adopted a model pruning approach (FairPrune), which mitigates unfairness issues by pruning unimportant model parameters and perform prediction using pruned-model.\\ \linebreak
\noindent
\textbf{Our Contributions:} 
The past researchers have utilized the teacher-student framework in mitigating bias issues~\cite{dhar2021distill,dong2023reliant,li2023dual,yue2024revisiting}. Most of these contributions mainly utilize unbiased teacher(s) or teacher networks with multi-domain information but suffer from accuracy loss while solving bias. Our novelty lies in making use of two biased teachers to solve bias issues and, at the same time, improve accuracy measures. To the best of our knowledge, we are the first to propose and validate the benefit of employing two biased teachers to mitigate biases and achieve fairness with accuracy uplift. We developed a multi-weighted loss comprised of biasing and debiasing components, which gives control towards the fairness of the model. We validated the effectiveness of the proposed approach against various SOTA methods and established a new benchmark on the fairness-accuracy trade-off using two widely used public dermatology datasets. Our extensive ablation study demonstrates the potential of the proposed approach in controlling the biases and debias directions of the model.

\section{Methodology} 
\subsection{Problem Formulation:} Consider a dataset, $D = \{x_i, y_i, k_i\}$, where $i \in \{1, \dots, n\}$ and $n$ represents the total number of samples present in the dataset. We denote the set of image samples as $X$, class labels as $C$, and sensitive attributes as $K$, such that $x \in X$, $y \in C$, and $k \in K = \{0,1\}$, where 0 and 1 denotes sensitive attribute type. We aim to build a fair classifier that improves the classification of input samples irrespective of their sensitive attribute type.
\begin{figure}
  \centering
  \includegraphics[trim=0cm 0cm 0cm 0cm, clip, width=0.99\textwidth]{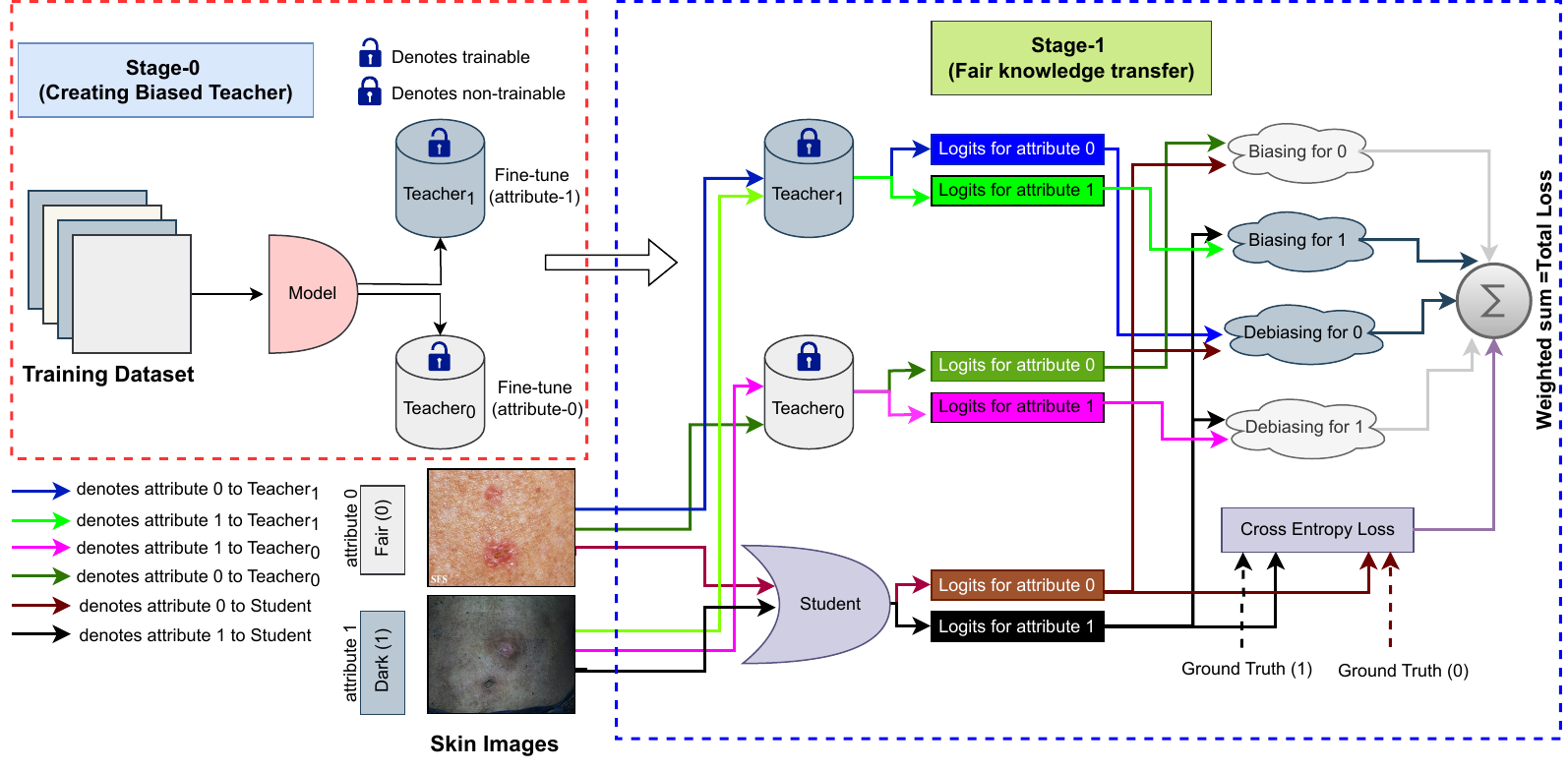}
  \caption{The schematic architecture diagram of the proposed approach. We have shown how individual loss terms use logits from biased teachers to perform fair knowledge transfer (better view in color).}
  \label{fig:proposed_approach}
\end{figure}

\subsection{Biasing and Debiasing Strategy:} To perform biasing and debiasing, we utilize `two biased teachers (say $T_0$ and $T_1$) and one student (say $S$)' based network. In our setup, $T_0$ and $T_1$ are biased to sensitive attributes 0 and 1, respectively. 

\noindent
If $X_0$ and $X_1$ represent the input samples for sensitive (protected) attribute type 0 and 1, respectively, then we can achieve biasing and debiasing using KL divergence-based loss functions~\cite{hinton2015distilling}: $KL_{bias_0}$, $KL_{bias_1}$ and $KL_{debias_0}$, $KL_{debias_1}$ respectively where 0, 1 denotes the sensitive attribute type for biasing or debiasing. We can formulate the biasing and debiasing loss terms using Eq.~\ref{eq:bias_0}, Eq.~\ref{eq:bias_1} and Eq.~\ref{eq:debias_0}, Eq.~\ref{eq:debias_1} respectively, where ${P_c}^f (X_K)$ denotes the softened predicted logit vector of class $c$ obtained through network $f_{\in \{S, T_0, T_1\}}$ when input sample belongs to sensitive attribute $K_{\in \{0, 1\}}$, (i.e., $X_K$) is passed through it. For any logit vector $z$, we compute its softened probability vector as shown in Eq.~\ref{eq:soft_pred}, where $z_i$ represents i-th value of the logit vector $z$, $C$ is the number of classes, and $\tau$ denotes the temperature coefficient. To perform the classification of the input sample (which may belong to bias group 0 or 1), we used cross-entropy loss (denoted as $L_{CE}$) as shown in Eq.~\ref{eq:CEL}, where $y_c$ and $z_c$ represents ground truth and logit vector for input sample.

\begin{equation}\label{eq:soft_pred}
      P_c = \frac{\exp(z_i/\tau)}{\sum_{c=1}^{C} \exp(z_c/\tau)}
\end{equation}

\begin{equation}\label{eq:CEL}
    L_{CE}(z, y) = \sum_{c=1}^C -y_c \log z_c 
\end{equation}

\begin{equation} \label{eq:bias_0}
    L_{bias_0} = \tau^2 \sum_{c=1}^{C} {{P_c}^{T_0}(X_{0})} \log \frac{{P_c}^{T_0}(X_{0})}{{P_c}^S(X_{0})}
\end{equation}

\begin{equation} \label{eq:bias_1}
    L_{bias_1} = \tau^2 \sum_{c=1}^{C} {{P_c}^{T_1}(X_{1})} \log \frac{{P_c}^{T_1}(X_{1})}{{P_c}^S(X_{1})}
\end{equation}

\begin{equation} \label{eq:debias_0}
    L_{debias_0} = \tau^2 \sum_{c=1}^{C} {{P_c}^{T_1}(X_{0})} \log \frac{{P_c}^{T_1}(X_{0})}{{P_c}^S(X_{0})}
\end{equation}

\begin{equation} \label{eq:debias_1}
    L_{debias_1} = \tau^2 \sum_{c=1}^{C} {{P_c}^{T_0}(X_{1})} \log \frac{{P_c}^{T_0}(X_{1})}{{P_c}^S(X_{1})}
\end{equation}

\newpage
\subsection{Achieving Fair Student Network:} Our proposed approach is inspired by the teacher-student-based knowledge distillation mechanism where the teacher (expert in a particular subject domain/ target task) distills their knowledge to a lightweight student network. To transfer fair knowledge from two biased teachers ($T_0$ and $T_1$) into a student network ($S$), we developed a weighted loss function that allows our student network to learn from two biased teachers through weighted biasing and debiasing in a balanced manner. For fair knowledge transfer, we utilized multi-loss composed of a weighted combination of biasing losses (expressed using Eq.~\ref{eq:bias_0}, Eq.~\ref{eq:bias_1}) and debiasing losses (expressed using Eq.~\ref{eq:debias_0}, Eq.~\ref{eq:debias_1}). The proposed loss function to ensure fairness in the student network is expressed in Eq.~\ref{eq:proposed_loss}. The schematic architecture diagram of the proposed approach has been illustrated in Fig.~\ref{fig:proposed_approach}.

\begin{equation} \label{eq:proposed_loss}
    L_{total} =\lambda~ L_{CE} + \alpha~ L_{bias_0} + \beta~ L_{bias_1} + \gamma~ L_{debias_0} + \delta~ L_{debias_1}  
\end{equation}
 where $\lambda, \alpha, \beta, \gamma, \delta$ are hyperparameters (decided experimentally) to adjust the weightage of different loss terms and hence control the extent of biasing and debiasing to ensure fair knowledge transfer. In Table~\ref{tab:ablation_fitz_isic}, we demonstrate the individual influence of biasing and debiasing loss terms at different weights.\\

\section{Experiments and Results}
\subsection{Datasets:} 
\begin{figure}[H]
    \centering
    \includegraphics[ width=0.99\linewidth, height=7cm, trim=0.7cm 70cm 9.5cm 1.2cm, clip]{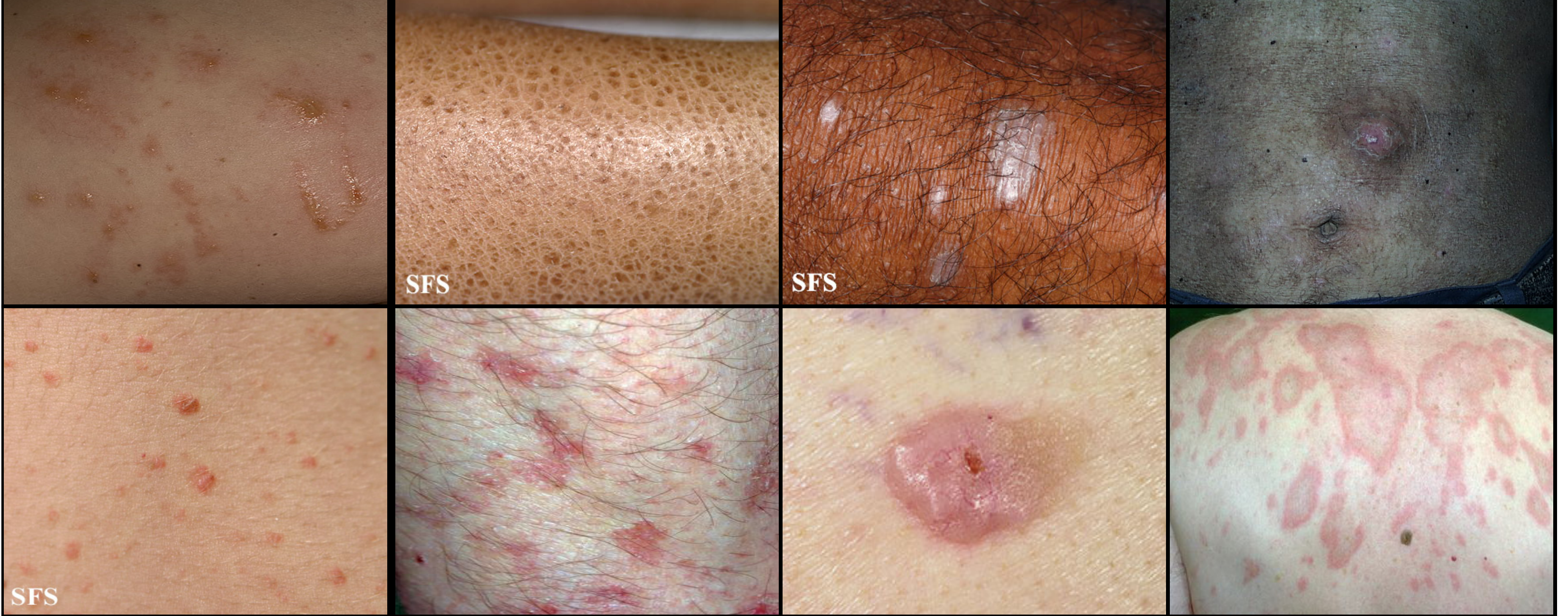}
    \caption{Examples of clinical images representing skin disease conditions in the Fitzpatrick-17k dataset. }
    \label{fig:examples_fitz_images}
\end{figure}

\begin{figure}[H]
    \centering
    \includegraphics[ width=0.99\linewidth, height=7cm, trim=0.5cm 78.8cm 32.9cm 1.1cm, clip]{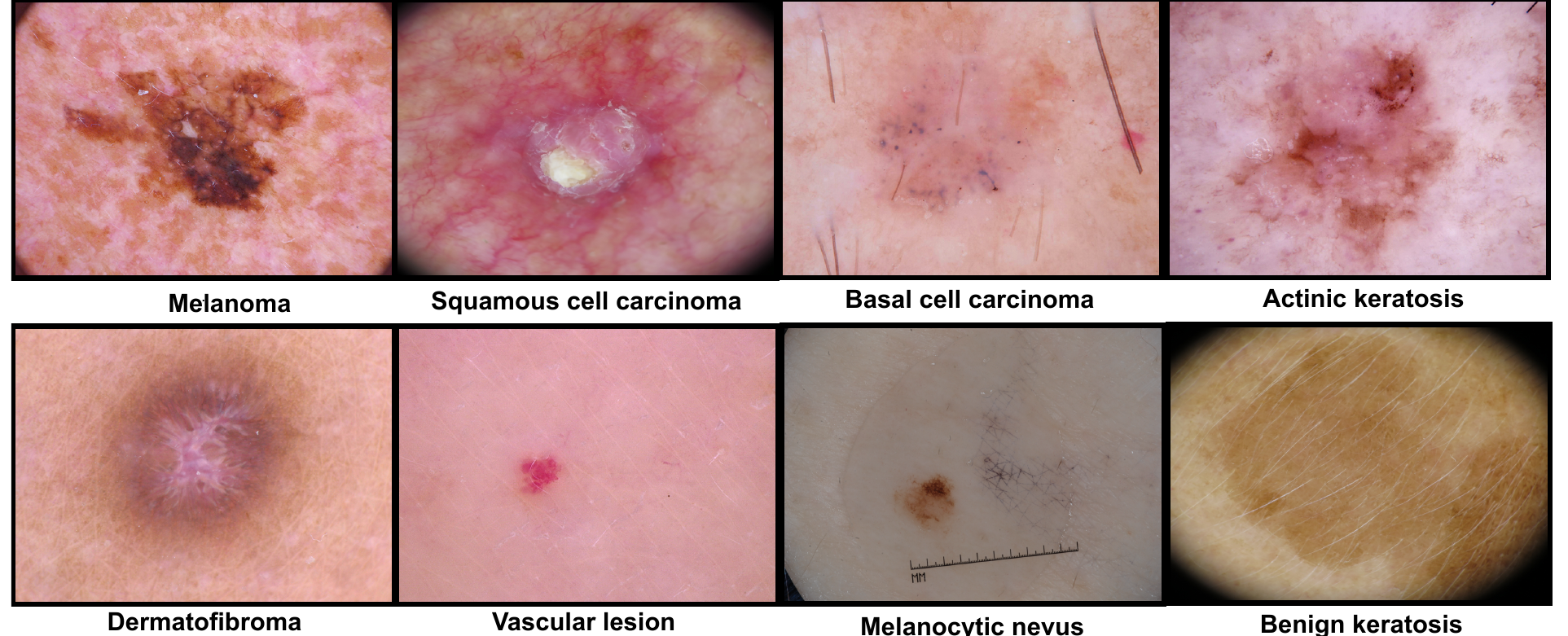}
    \caption{Examples of dermoscopic images representing skin lesion classes in the ISIC-2019 dataset.}
    \label{fig:examples_isic_images}
\end{figure}

We used two publicly available dermatology datasets, i.e., Fitzpatrick-17k and ISIC-2019 to evaluate our approach. The Fitzpatrick-17k dataset contains 16,577 skin condition images comprising 114 classes across six skin tone groups (determined by the Fitzpatrick scale from 1 to 6). We considered images with a Fitzpatrick scale of 1 to 3 as fair and the remaining as dark, which leads to two skin groups (fair and dark) as skin tone bias. The ISIC-2019 dataset contains 25,331 dermatology skin lesion images comprised of 9 diagnostic categories. Due to the unavailability of skin tone information, we consider available gender information and use female/male information as sensitive (protected) attributes. We provided the images for Fitzpatrick-17k dataset and ISIC-2019 dataset in Fig.~\ref{fig:examples_fitz_images} and Fig~\ref{fig:examples_isic_images} respectively.\\ 

\newpage
\subsection{Implementation Details:} The proposed approach has been implemented using the PyTorch and train/ test with a 16GB NVIDIA V100 GPU. To establish fair performance with the recent state-of-the-art (SOTA) method (i.e., FairPrune~\cite{wu2022fairprune}), we used VGG-11 with Fitzpatrick-17k and ResNet-18 with ISIC-2019 initialized with pre-trained ImageNet weights. We implemented our models with 200 epochs, batch size 128, SGD optimizer, learning rate 0.01, and used the same data split as FairPrune. We used VGG-13 and ResNet-50 as Teacher networks for Fitzpatrick and ISIC-2019 datasets. The choice of teacher network variant is to promote model generalization and to become heavily biased towards the different sensitive attributes present in the given dataset.\\To obtain fair and dark teachers for the Fitzpatrick-17k dataset, we firstly, we train VGG-13 on a train split dataset. To build biased teachers for the fair bias group and dark bias group, we finetune the model on fair samples and dark samples present in the given train split, respectively. For fair teacher, we obtained F1-score of 0.5475 and 0.5651 for dark teacher. Similarly, for ISIC-2019, male and female teachers were obtained by finetuning the ResNet-50 model (using train split of ISIC-2019) on male and female samples and achieved an F1-score of 76.02 and 0.7704, respectively. We experimentally decided $\tau$ = 5 for the Fitzpatrick dataset and $\tau$ = 4 for the ISIC-2019 dataset. Similarly, other hyper-parameters ($\lambda, \alpha$, $\beta$, $\gamma$, and $\delta$) were decided experimentally as \{1, 0.99, 0.001, 0.99, 0.01\} and \{1, 0.2, 0.001, 0.999, 0\} for Fitzpatrick-17k, ISIC-2019 dataset respectively.\\

\subsection{Evaluation Measures:} For a fair evaluation of our approach on different datasets, we followed the same evaluation measure as followed by FairPrune. We compute Precision, Recall, and F1-score to evaluate accuracy measures across sensitive groups. For fairness evaluation, we used multi-class equalized opportunity (Eopp0, Eopp1) and equalized odds (Eodd) as proposed in~\cite{hardt2016equality}. We denote True Positive, True Negative, False Positive, and False Negative for class $c$ with sensitive group $k=\{0, 1\}$ as $TP_c^k$, $TN_c^k$, $FP_c^k$, and $FN_c^k$ respectively. Then, we can formally define Eopp0, Eopp1, and Eodd using Eq.~\ref{eq:Eopp0}, Eq.~\ref{eq:Eopp1} and Eq.~\ref{eq:Eodd} respectively.\\  
\begin{equation}\label{eq:Eopp0}
    Eopp0 = \sum_{c=1}^{C} |TNR_c^1 - TNR_c^0|
\end{equation}

\begin{equation}\label{eq:Eopp1}
Eopp1 = \sum_{c=1}^{C} |TPR_c^1 - TPR_c^0|
\end{equation}

\begin{equation}\label{eq:Eodd}
    Eodd = \sum_{c=1}^{C} |TPR_c^1 - TPR_c^0 + FPR_c^1 - FPR_c^0|
\end{equation}

where $TPR_c^k$, $TNR_c^k$, and $FPR_c^k$ denote the True Positive Rate, True Negative Rate, and False Positive Rate for sensitive group $k$ belonging to class $c$. We compute $TPR_c^k$, $TNR_c^k$, and $FPR_c^k$ using Eq.~\ref{eq:TPR}, Eq.~\ref{eq:TNR}, and Eq.~\ref{eq:FPR}, respectively.

\begin{equation}\label{eq:TPR}
     TPR_c^k = \frac{TP_c^k}{TP_c^k+FN_c^k}
\end{equation}

\begin{equation}\label{eq:TNR}
    TNR_c^k = \frac{TN_c^k}{TN_c^k+FP_c^k}
\end{equation}

\begin{equation}\label{eq:FPR}
     FPR_c^k = \frac{FP_c^k}{FP_c^k+TN_c^k}
\end{equation}

\subsection{Comparison with State-of-the-Arts:} We utilize Fitzpatrick-17k and ISIC-2019 datasets to compare the proposed approach on various baselines and SOTA approaches, including VGG-11, ResNet-18, AdvConf~\cite{alvi2018turning}, AdvRev~\cite{zhang2018mitigating}, DomainIndep~\cite{wang2020towards}, HSIC~\cite{quadrianto2019discovering}, and MFD~\cite{jung2021fair}, FairPrune~\cite{wu2022fairprune}. We summarise the SOTA comparison on Fitzpatrick-17k and ISIC-2019 in Table~\ref{tab:sota_fitz} and Table~\ref{tab:sota_isic}, respectively.\\

\noindent
\textbf{A better solution ensures Fairness while improving prediction:} It is important to note that various methods, including the baseline, achieved predictive accuracy around 50\% for the Fitzpatrick-17k dataset. However, they struggled to maintain fairness, resulting in poor fairness metrics (i.e., higher values for Eopp0, Eopp1, and Eodd). Similarly, for the ISIC-2019 dataset, the baseline model and other approaches achieved an F-score of around 73\%, but they also failed to maintain fairness, showing poor fairness metrics. For both datasets, our proposed approach not only improved prediction across bias groups without compromising fairness metrics but also enhanced fairness (i.e., showing lower values for Eopp0, Eopp1, and Eodd).\\
It is crucial to note that fairness metrics differ from predictive measures in terms of relative numerical significance. A small change in the F-score might not be significant, but even a slight improvement in fairness metrics holds significant importance. This is because even minor disparities in fairness metrics can have significant real-world consequences, such as reinforcing biases or perpetuating discrimination. For instance, we can see that the baseline models achieved comparable average F-scores to state-of-the-art approaches, but they failed to maintain fairness.

\subsection{Time analysis:}
It is important to note that during model training, i.e., training of student network and distilling the knowledge from the teacher networks, we keep the teacher networks frozen, i.e., the teacher networks remain non-trainable. 
After training the student model, we only require the student model to perform testing on the input image, i.e., during test time, we do not require access to the teacher networks. Considering the fact that model training is required only once and practical usage only requires prediction using the trained student model, the proposed approach supports the practical utility of our student network, which contains the distilled fair knowledge by using the multiple-teacher student paradigm. Overall, our model took 2 hours 6 minutes to train and 2.57 seconds in total for test time prediction for the entire test dataset. 

\subsection{Ablation Study:} We conducted an ablation study to analyze the effect of individual loss terms in conjunction with cross-entropy loss at the different weights (=\{0.6, 0.8, 1.0\}). Our ablation study validates the potential of biasing terms ($L_{bias_0}$, $L_{bias_1}$) and debiasing terms ($L_{debias_0}$, $L_{debias_1}$) and therefore increased the f1-score during biasing and reduced during debiasing, which is essential in maintaining the balance among bias groups to fairness during classification. We also observed that an increase in weight helps to amplify the effect of loss terms. The summary of the ablation study to demonstrate the utility of biasing and debiasing to ensure fairness with performance has been provided in Table~\ref{tab:ablation_fitz_isic}.

\setlength{\tabcolsep}{4pt}
\begin{table}[H]
    \centering
  \caption{Summary of ablation study to analyze the potential of biasing and debiasing terms at different weights. Here, F(0) and F(1) denote the F1-score obtained by the model for sensitive groups 0 and 1, respectively. Here, bold and underlined entries denote increment and decrement in the F1-score, respectively.}
    \label{tab:ablation_fitz_isic}
    \fontsize{10}{11}\selectfont 
    \begin{tabular}{@{}c|ccccccccc@{}}
        \midrule
        \multirow{2}{*}{\textbf{Weight}} & \multicolumn{5}{c}{\textbf{Loss Terms}} & \multicolumn{2}{c}{\textbf{Fitzpatrick-17k}} & \multicolumn{2}{c}{\textbf{ISIC-2019}} \\
        \cline{2-10}
        & $L_{CE}$ & $L_{bias_0}$ & $L_{bias_1}$ & $L_{debias_0}$ & $L_{debias_1}$ & \textbf{F(0)} & \textbf{F(1)} & \textbf{F(0)} & \textbf{F(1)} \\
        \midrule
        0 & \ding{51} & \ding{55} & \ding{55} & \ding{55} & \ding{55} & 0.4730 & 0.5460 & 0.7230 & 0.7460 \\
       
       \midrule
       {0.6} & \ding{51} & \ding{51} & \ding{55} & \ding{55} & \ding{55} & \textbf{0.5151} & 0.5273 & \textbf{0.7302} & 0.7431 \\
        {0.8} & \ding{51} & \ding{51} & \ding{55} & \ding{55} & \ding{55} &\textbf{0.5159}  & 0.5184 & \textbf{0.7311} & 0.7374 \\
        {1.0} & \ding{51} & \ding{51} & \ding{55} & \ding{55} & \ding{55} & \textbf{ 0.5194} &0.5468 & \textbf{0.7352} & 0.7245 \\
        \midrule
        
        {0.6}& \ding{51} & \ding{55} & \ding{51} & \ding{55} & \ding{55} & 0.4881 & \textbf{0.5292} & 0.7185 & \textbf{0.7399} \\
         {0.8}& \ding{51} & \ding{55} & \ding{51} & \ding{55} & \ding{55} & 0.5028 & \textbf{0.5486} & 0.7116 & \textbf{0.7442} \\
          {1.0}& \ding{51} & \ding{55} & \ding{51} & \ding{55} & \ding{55} & 0.5051  &\textbf{0.5634} & 0.7089 & \textbf{0.7518} \\
         \midrule
        {0.6}& \ding{51} & \ding{55} & \ding{55} & \ding{51} & \ding{55} &\underline{0.4673}  & 0.5267 & \underline{0.7264} & 0.7485 \\
        {0.8}& \ding{51} & \ding{55} & \ding{55} & \ding{51} & \ding{55} &  \underline{0.4557}  &0.5245& \underline{0.7175} & 0.7480 \\
        {1.0}& \ding{51} & \ding{55} & \ding{55} & \ding{51} & \ding{55} & \underline{0.4546} & 0.5159 & \underline{0.7099} & 0.7232 \\
        \midrule
         {0.6}& \ding{51} & \ding{55} & \ding{55} & \ding{55} & \ding{51} & 0.4994 & \underline{0.5321} & 0.7129 & \underline{0.7374} \\
        {0.8}& \ding{51} & \ding{55} & \ding{55} & \ding{55} & \ding{51} & 0.5016  &\underline{0.5252} & 0.7174 & \underline{0.7226} \\
        {1.0}& \ding{51} & \ding{55} & \ding{55} & \ding{55} & \ding{51} &  0.4957 &\underline{0.5107} & 0.7216 &\underline{0.7199}  \\
        \midrule
        \multirow{1}{*}{\textbf{\textit{Proposed}}} & \ding{51} & \ding{51} & \ding{51} & \ding{51} & \ding{51} & \textit{0.5207}  & \textit{0.5263} & \textit{0.7388} & \textit{0.7400} \\ \midrule
    \end{tabular}
\end{table}

\subsection{Qualitative Results:}
\begin{figure}[H]
  \centering
  \begin{subfigure}[b]{0.85\textwidth}
    \centering
    \includegraphics[trim=0cm 0cm 0cm 0cm, clip, width=\textwidth, height = 8cm]{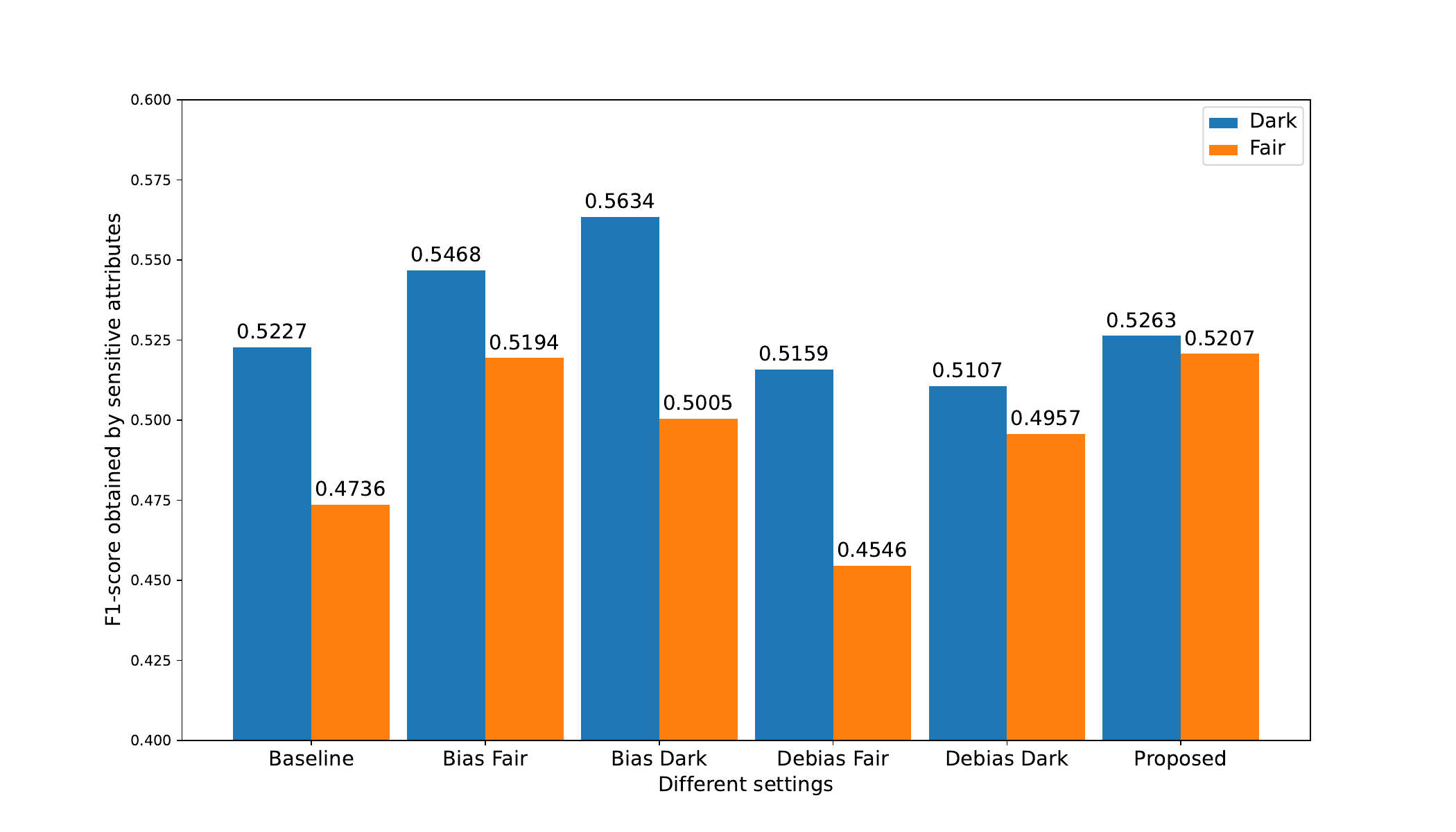}
    \caption{On Fitzpatrick-17k dataset}
    \label{subfig:BAR_PLOT_FITZ_1}
  \end{subfigure}\\
  \begin{subfigure}[b]{0.85\textwidth}
    \centering
    \includegraphics[trim=0cm 0cm 0cm 0cm, clip, width=\textwidth, height = 8cm]{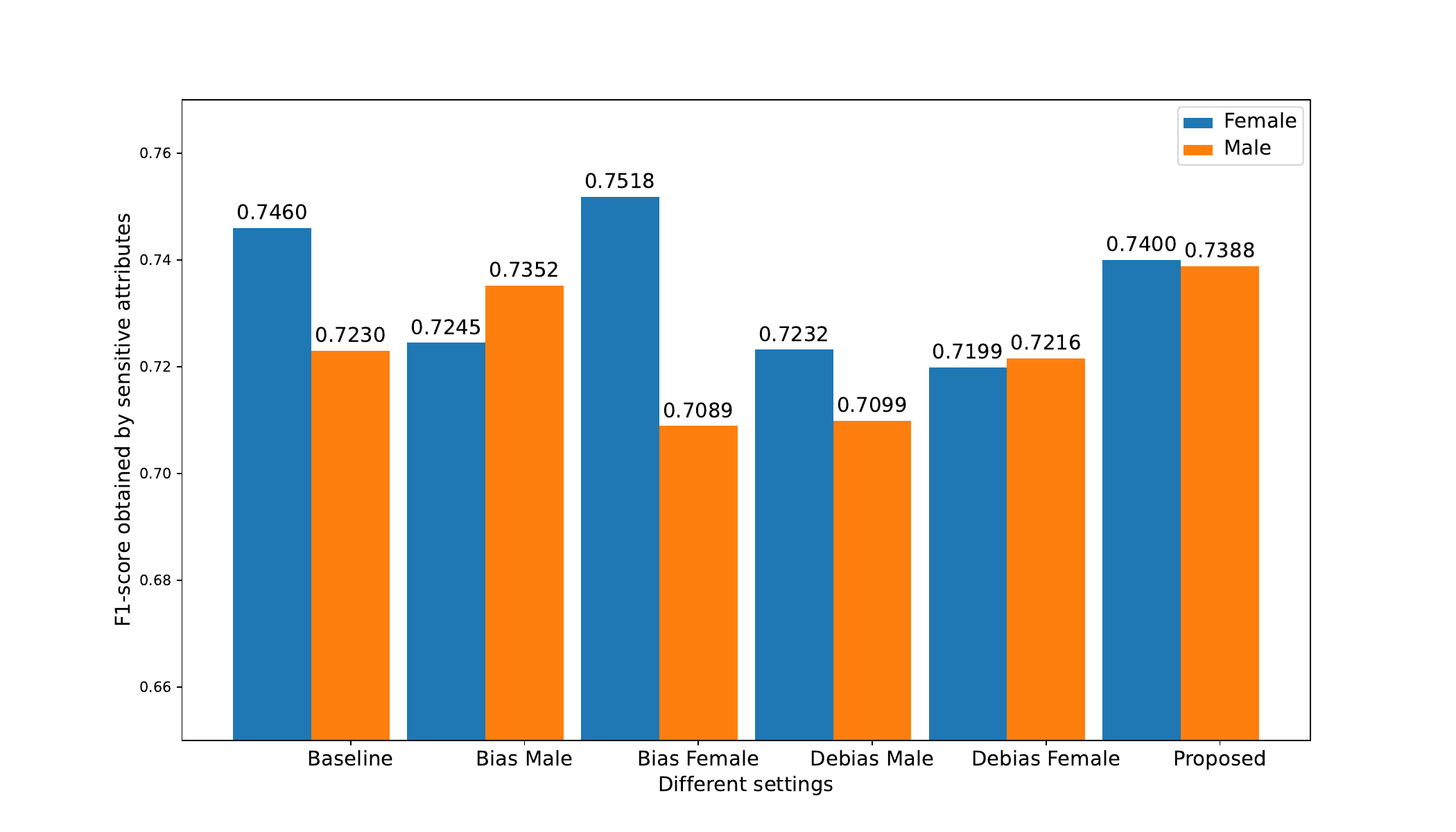}
    \caption{On ISIC-2019 dataset}
    \label{subfig:BAR_PLOT_ISIC_1}
  \end{subfigure}\\
\caption{Fairness comparison using bar plots (a), (b). We can see that the proposed approach ensures fairness across bias groups without sacrificing performance. Here, we used weight=1 for Bias Fair, Bias  Dark, Debias Fair, Debias Dark.}
  \label{fig:BAR_PLOTS}
\end{figure}

\begin{figure}[H]
  \centering
  \begin{subfigure}[b]{0.85\textwidth}
    \centering
    \includegraphics[trim=0cm 0cm 0cm 0cm, clip, width=\textwidth, height=8cm]{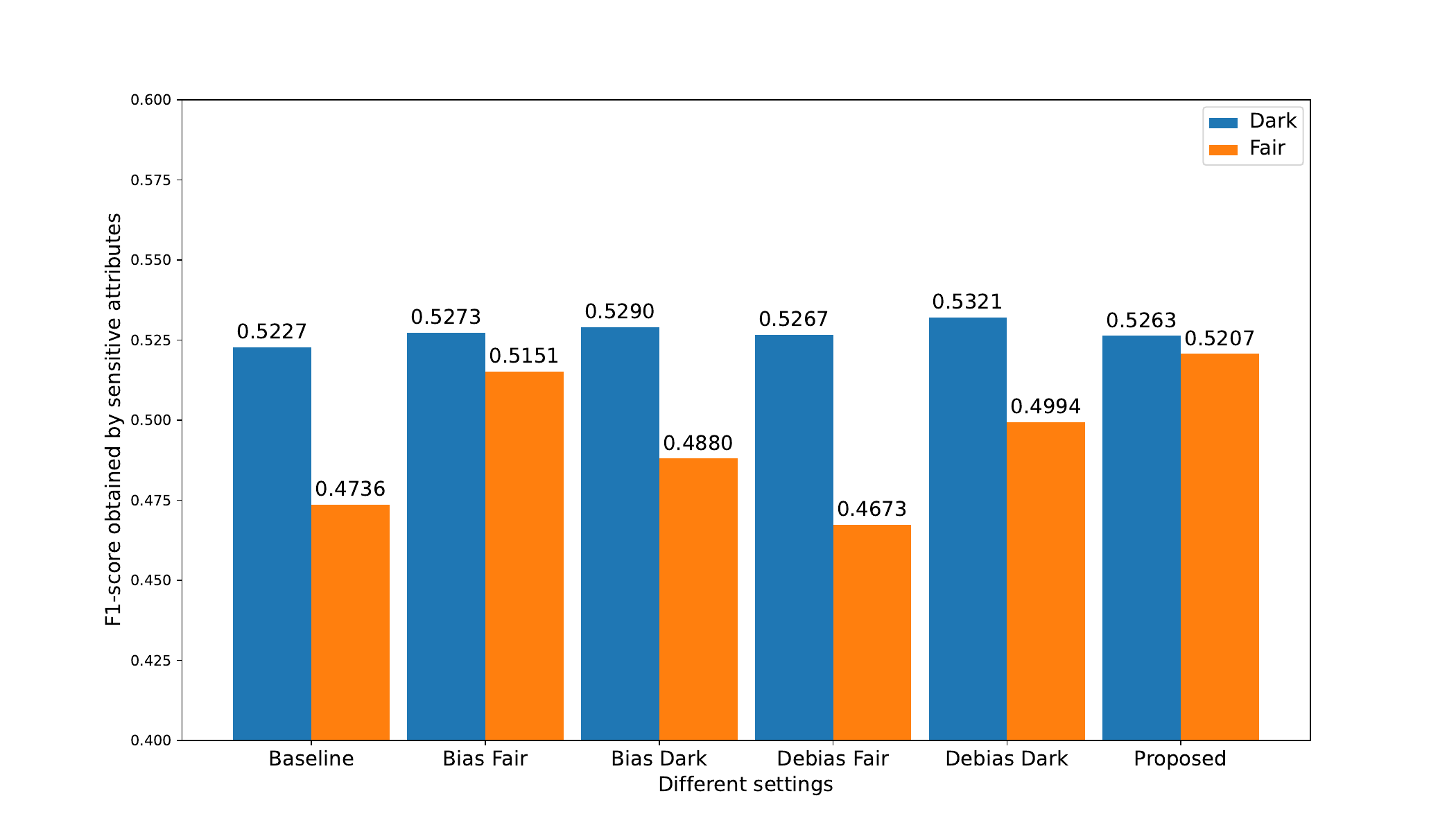}
    \caption{On Fitzpatrick-17k dataset}\label{subfig:BAR_PLOT_FITZ_0.6}
  \end{subfigure}
  \hfill
  \begin{subfigure}[b]{0.85\textwidth}
    \centering
    \includegraphics[trim=0cm 0cm 0cm 0cm, clip, width=\textwidth, height=8cm]{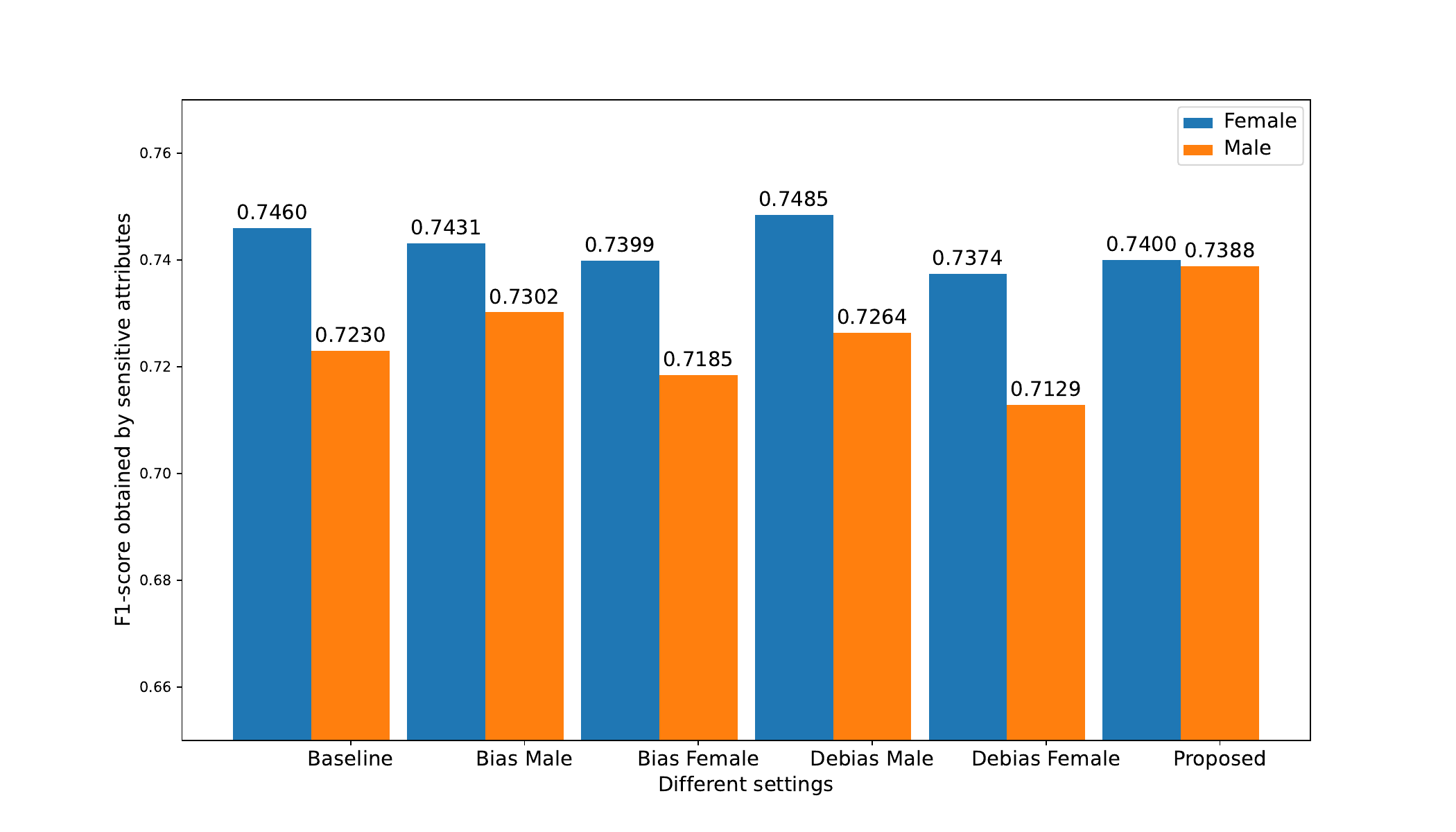}
    \caption{On ISIC-2019 dataset}\label{subfig:BAR_PLOT_ISIC_0.6}
  \end{subfigure}
  \caption{The bar plot obtained while employing different biasing and debiasing loss components at weight coefficient=0.6. We can observe that the proposed approach provides improvement over the baseline while ensuring fairness.}\label{fig:BAR_PLOTS_FITZ_ISIC_0.6}
\end{figure}

\begin{figure}[H]
  \centering
  \begin{subfigure}[b]{0.85\textwidth}
    \centering
    \includegraphics[trim=0cm 0cm 0cm 0cm, clip, width=\textwidth, height=8cm]{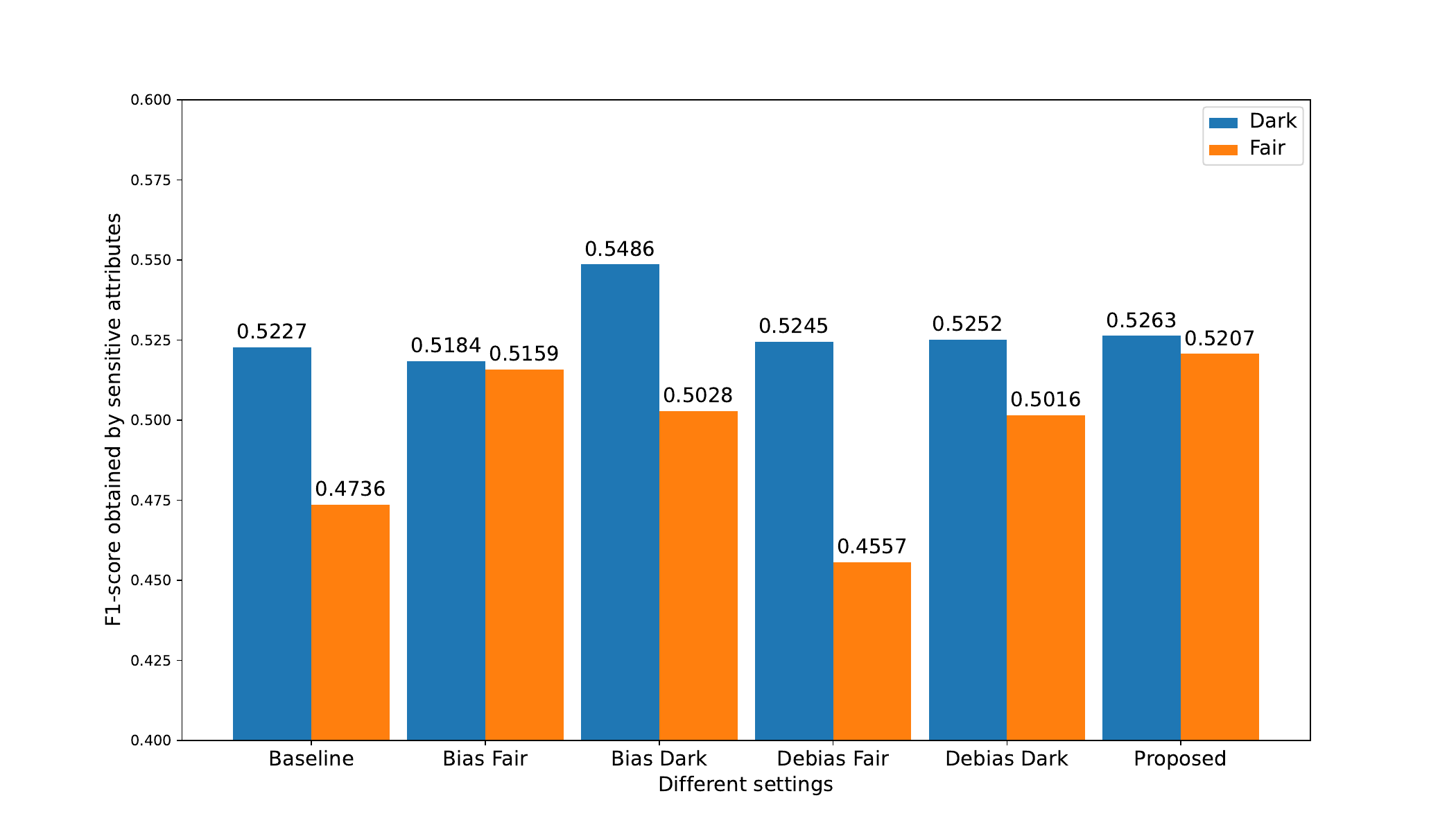}
    \caption{On Fitzpatrick-17k dataset}\label{subfig:BAR_PLOT_FITZ_0.8}
  \end{subfigure}
  \hfill
  \begin{subfigure}[b]{0.85\textwidth}
    \centering
    \includegraphics[trim=0cm 0cm 0cm 0cm, clip, width=\textwidth, height=8cm]{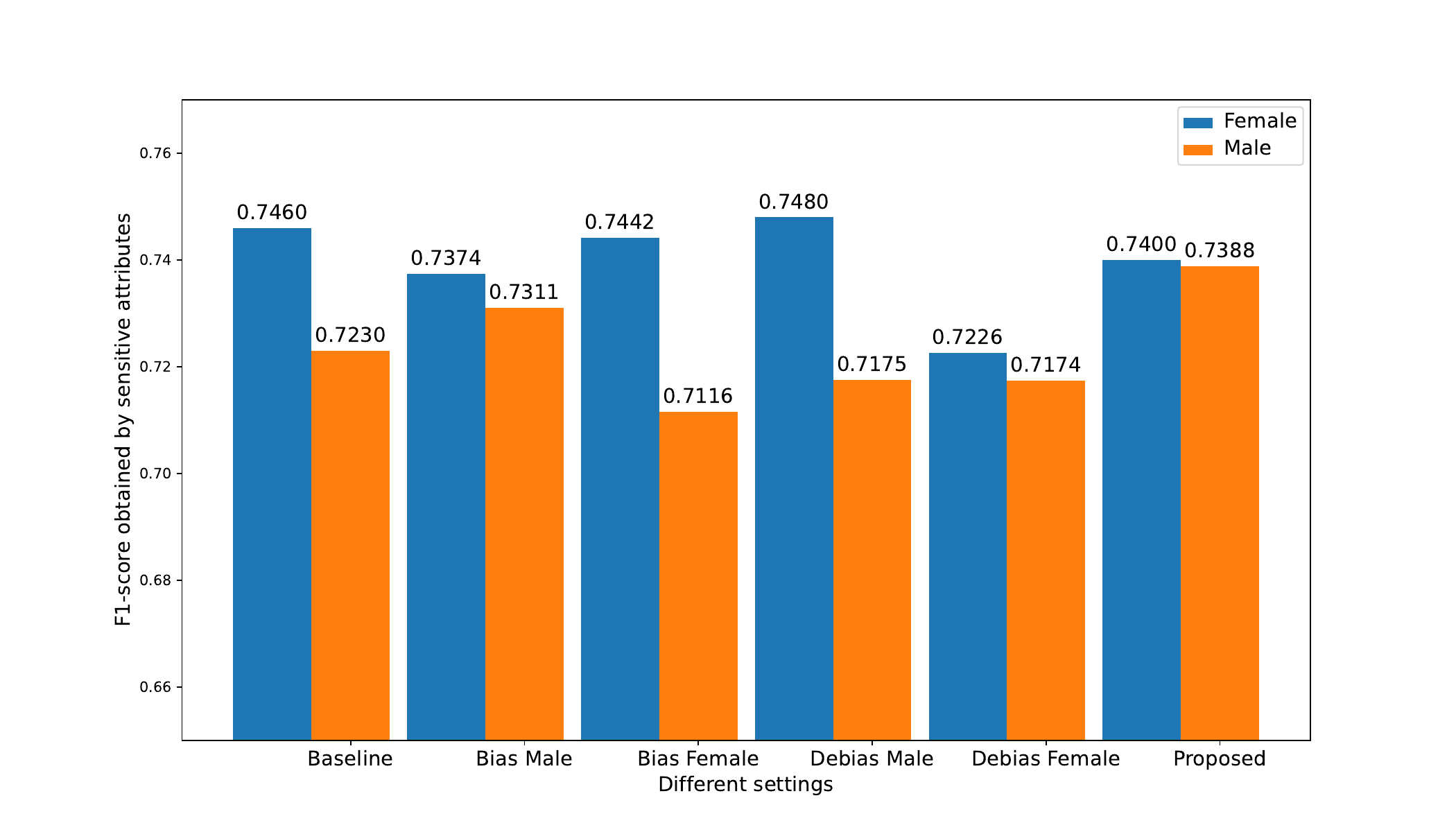}
    \caption{On ISIC-2019 dataset}\label{subfig:BAR_PLOT_ISIC_0.8}
  \end{subfigure}
  \caption{The bar plot obtained while employing different biasing and debiasing loss components at weight coefficient=0.8. We can observe that the proposed approach provides improvement over the baseline while ensuring fairness.}\label{fig:BAR_PLOTS_FITZ_ISIC_0.8}
\end{figure}

\begin{figure}[H]
  \centering
  
  \begin{subfigure}[b]{0.85\textwidth}
    \centering
    \includegraphics[trim=2.2cm 1.7cm 2.5cm 2.3cm, clip, width=\textwidth, height = 8cm]{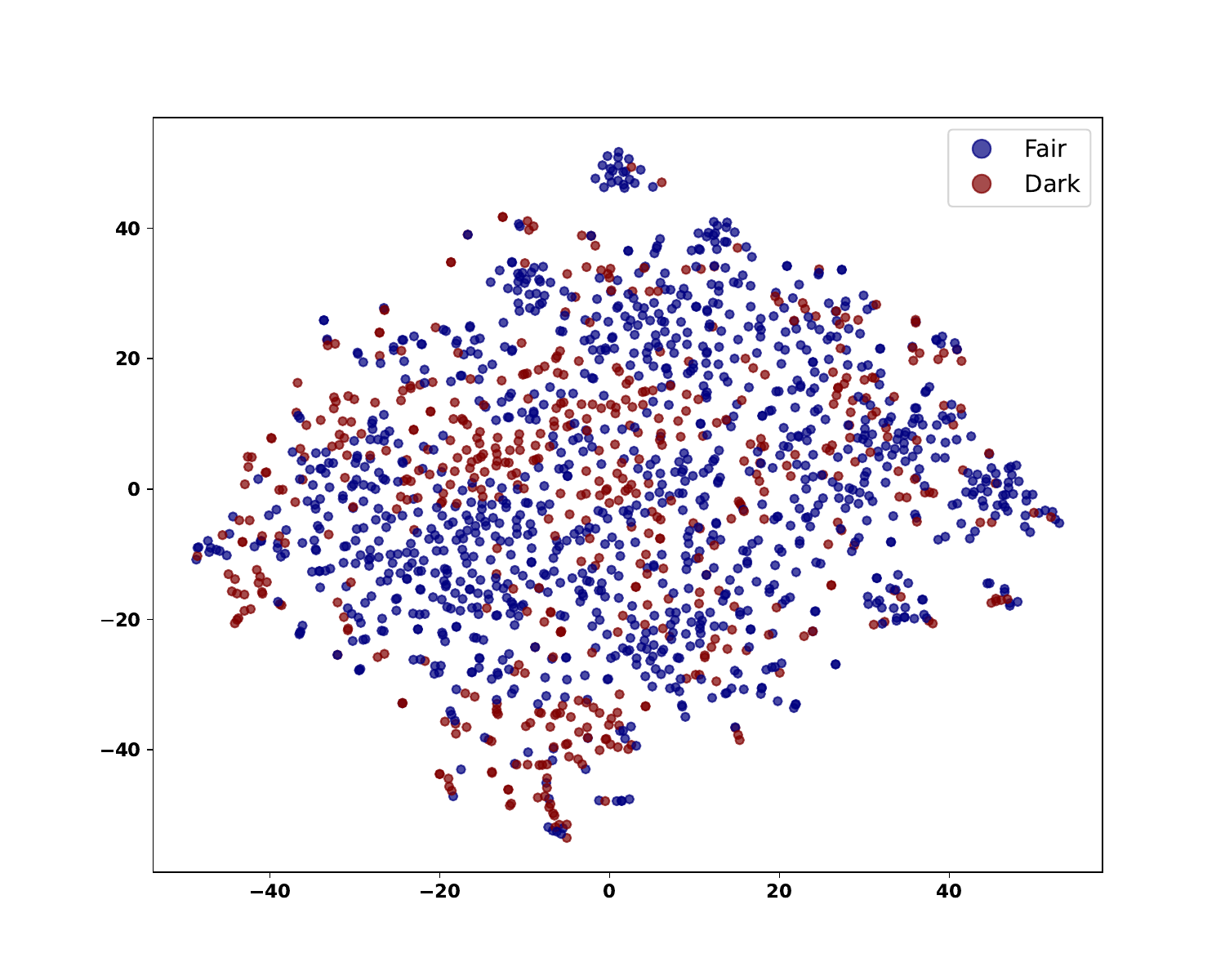}
    \caption{Baseline model (unfair)}
    \label{subfig:TSNE_PLOT_FITZ}
  \end{subfigure}\\
  \begin{subfigure}[b]{0.85\textwidth}
    \centering
    \includegraphics[trim=2.2cm 1.7cm 2.5cm 2.3cm, clip, width=\textwidth, height = 8cm]{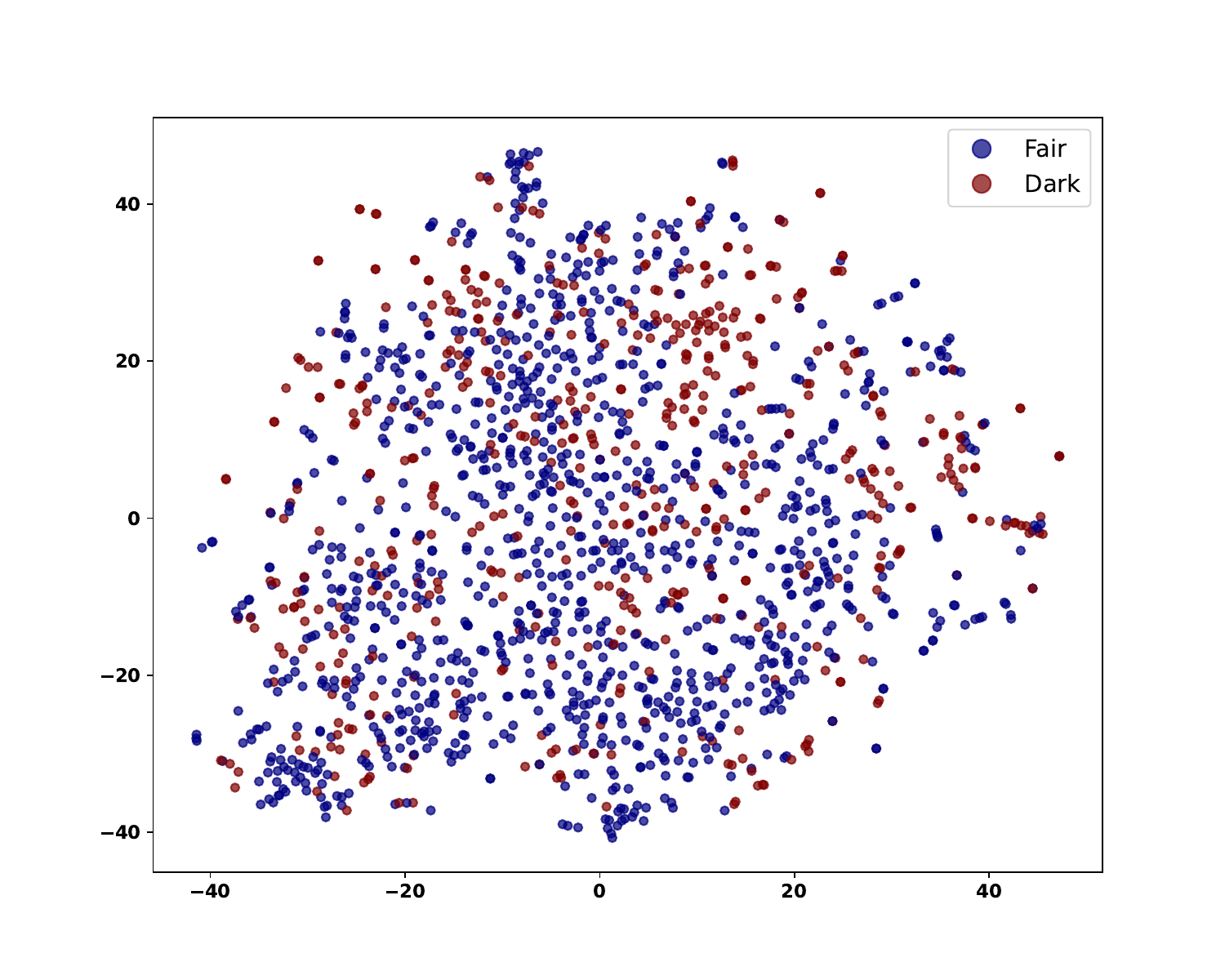}
    \caption{Proposed model (fair)}
    \label{subfig:TSNE_PLOT_ISIC}
  \end{subfigure}
\caption{Fairness comparison using t-SNE with respect to sensitive attributes in plots (a) and (b). We can see that the proposed approach contains a wider spread among data points which means the data points are less entangled in terms of sensitive attributes, i.e., the proposed model is fair in comparison to the baseline model.}
  \label{fig:TSNE_PLOTS}
\end{figure}

We demonstrate the effectiveness of the proposed approach qualitatively by comparing the obtained performance at weight = \{1.0, 0.6, 0.8\} for Fitzpatrick-17k and ISIC-2019 datasets under different settings in Fig.~\ref{fig:BAR_PLOTS}, Fig.~\ref{fig:BAR_PLOTS_FITZ_ISIC_0.6} and Fig.~\ref{fig:BAR_PLOTS_FITZ_ISIC_0.8} respectively. Through bar plots, we proved the utility of biasing and debiasing terms to control the F1-score of the baseline model for available bias groups and ensure fair knowledge transfer while increasing accuracy. Using Fig.~\ref{fig:TSNE_PLOTS}, we provided the t-SNE visualization based feature space visualization where more spread in dots for a fair model affirms that our proposed model is not finding similarity among samples based on sensitive attributes and thus is not holding them closer as found in the baseline model.

\setlength{\tabcolsep}{2pt}
\begin{table}[H]
    \centering
    \caption{Comparison with SOTA approaches on Fitzpatrick-17k dataset. In the proposed approach, we used VGG-11 in the proposed network for fair comparative analysis.}
    \label{tab:sota_fitz}
    \fontsize{9}{10}\selectfont
    \begin{tabular}{@{}lcccccccc@{}}
     \midrule
         \textbf{} & \textbf{} & \multicolumn{3}{c}{\textbf{Accuracy}$\uparrow$} & \multicolumn{3}{c}{\textbf{Fairness}$\downarrow$} \\
         \midrule
        \textbf{Method} & \textbf{Bias Group} & \textbf{Precision (\%)} & \textbf{Recall(\%)}& \textbf{F-score(\%)} & \textbf{Eopp0}$\downarrow$ & \textbf{Eopp1}$\downarrow$ & \textbf{Eodd}$\downarrow$ \\
        \midrule
        \multirow{4}{*}{VGG-11} & Dark & 56.3 & 58.1 & 54.6 & \multirow{4}{*}{0.0013} & \multirow{4}{*}{0.361} & \multirow{4}{*}{0.182} \\
         & Light & 48.2 & 49.5 & 47.3 &  &  &  \\
         & Avg.$\uparrow$ & 52.3 & 53.8 & 51.0 &  &  &  \\
         & Diff.$\downarrow$ & 8.1 & 8.6 & 7.3 &  &  &  \\
       \midrule
        \multirow{4}{*}{AdvConf} & Dark & 50.6 & 56.2 & 50.6 & \multirow{4}{*}{0.0011} & \multirow{4}{*}{0.339} & \multirow{4}{*}{0.169} \\
         & Light & 42.7 & 46.4 & 42.6 &  &  &  \\
         & Avg.$\uparrow$ & 46.7 & 51.3 & 46.6 &  &  &  \\
         & Diff.$\downarrow$ & 7.9 & 9.8 & 8.0 &  &  &  \\
         \midrule
        \multirow{4}{*}{AdvRev} & Dark & 51.4 & 54.5 & 50.3 & \multirow{4}{*}{0.0011} & \multirow{4}{*}{0.334} & \multirow{4}{*}{0.166} \\
         & Light & 48.9 & 46.9 & 45.7 &  &  &  \\
         & Avg.$\uparrow$ & 50.2 & 50.7 & 48.0 &  &  &  \\
         & Diff.$\downarrow$ & 2.5 & 7.6 & 4.6 &  &  &  \\
        \midrule
        \multirow{4}{*}{DomainIndep} & Dark & 55.9 & 54.0 & 53.0 & \multirow{4}{*}{0.0012} & \multirow{4}{*}{0.323} & \multirow{4}{*}{0.161} \\
         & Light & 54.1 & 52.9 & 51.2 &  &  &  \\
         & Avg.$\uparrow$ & 55.0 & 53.4 & 52.1 &  &  &  \\
         & Diff.$\downarrow$ & 1.8 & 1.0 & 1.8 &  &  &  \\
       \midrule
        \multirow{4}{*}{HSIC} & Dark & 54.8 & 52.2 & 51.3 & \multirow{4}{*}{0.0013} & \multirow{4}{*}{0.331} & \multirow{4}{*}{0.166} \\
         & Light & 51.3 & 50.6 & 48.6 &  &  &  \\
         & Avg.$\uparrow$ & 53.0 & 51.5 & 50.0 &  &  &  \\
         & Diff.$\downarrow$ & 4.0 & 1.8 & 2.9 &  &  &  \\
       \midrule
        \multirow{4}{*}{MFD} & Dark & 51.4 & 54.5 & 50.3 & \multirow{4}{*}{0.0011} & \multirow{4}{*}{0.334} & \multirow{4}{*}{0.166} \\
         & Light & 48.9 & 46.9 & 45.7 &  &  &  \\
         & Avg.$\uparrow$ & 50.2 & 50.7 & 48.0 &  &  &  \\
         & Diff.$\downarrow$ & 2.5 & 7.6 & 4.6 &  &  &  \\
         \midrule
        \multirow{4}{*}{FairPrune} & Dark & 56.7 & 51.9 & 50.7 & \multirow{4}{*}{\textbf{0.0008}} & \multirow{4}{*}{0.33} & \multirow{4}{*}{0.165} \\
         & Light & 49.6 & 47.7 & 45.9 &  &  &  \\
         & Avg.$\uparrow$ & 53.1 & 49.8 & 48.3 &  &  &  \\
         & Diff.$\downarrow$ & 7.1 & 4.2 & 4.8 &  &  &  \\
       \midrule
        \multirow{4}{*}{Proposed} & Dark & 56.96 & 53.20 & 52.63 & \multirow{4}{*}{{0.0012}} & \multirow{4}{*}{\textbf{0.2984}} & \multirow{4}{*}{\textbf{0.1491}} \\
         & Light & 54.38 & 54.22 & 52.07 &  &  &  \\
         & Avg.$\uparrow$ & 55.67 & 53.71 & \textbf{52.35} &  &  &  \\
         & Diff.$\downarrow$ & 2.58 & 1.02 & \textbf{0.56} &  &  &  \\
        \bottomrule
    \end{tabular} 
\end{table}

\begin{table}[H]
    \centering
    \caption{Comparison with SOTA approaches on ISIC-2019 Dataset. In the proposed approach, we used ResNet-18 in the proposed network for fair comparative analysis.}
    \label{tab:sota_isic}
    \fontsize{9}{10}\selectfont
    \begin{tabular}{@{}llcccccc@{}}
        \midrule
         \textbf{} & \textbf{} & \multicolumn{3}{c}{\textbf{Accuracy}$\uparrow$} & \multicolumn{3}{c}{\textbf{Fairness}$\downarrow$} \\
         \midrule
        \textbf{Method} & \textbf{Bias Group} & \textbf{Precision(\%)} & \textbf{Recall(\%)}& \textbf{F-score(\%)} & \textbf{Eopp0}$\downarrow$ & \textbf{Eopp1}$\downarrow$ & \textbf{Eodd}$\downarrow$ \\
        \midrule
        \multirow{4}{*}{ResNet-18} & Female & 79.3 & 72.1 & 74.6 & \multirow{4}{*}{0.006} & \multirow{4}{*}{0.044} & \multirow{4}{*}{0.022} \\
        & Male & 73.1 & 72.5 & 72.3 & & & \\
        & Avg.$\uparrow$ & 76.2 & 72.3 & 73.5 & & & \\
        & Diff.$\downarrow$ & 6.3 & 0.4 & 2.3 & & & \\
      \midrule
        \multirow{4}{*}{AdvConf} & Female & 75.5 & 73.8 & 74.1 & \multirow{4}{*}{0.008} & \multirow{4}{*}{0.07} & \multirow{4}{*}{0.037} \\
        & Male & 71.0 & 75.7 & 73.1 & & & \\
        & Avg.$\uparrow$ & 73.3 & 74.7 & 73.6 & & & \\
        & Diff.$\downarrow$ & 4.5 & 2.0 & 1.0 & & & \\
        \midrule
        \multirow{4}{*}{AdvRev} & Female & 77.8 & 68.3 & 71.6 & \multirow{4}{*}{0.007} & \multirow{4}{*}{0.059} & \multirow{4}{*}{0.033} \\
        & Male & 77.3 & 70.6 & 72.9 & & & \\
        & Avg.$\uparrow$ & 77.5 & 69.4 & 72.3 & & & \\
        & Diff.$\downarrow$ & 0.6 & 2.3 & 1.4 & & & \\
      \midrule
        \multirow{4}{*}{DomainIndep} & Female & 72.9 & 74.7 & 73.4 & \multirow{4}{*}{0.01} &\multirow{4}{*} {0.086} &\multirow{4}{*} {0.042} \\
        & Male & 72.5 & 69.4 & 70.2 & & & \\
        & Avg.$\uparrow$ & 72.7 & 72.1 & 71.8 & & & \\
        & Diff.$\downarrow$ & 0.4 & 5.3 & 3.1 & & & \\
       \midrule
        \multirow{4}{*}{HSIC} & Female & 74.4 & 66.0 & 69.6 & \multirow{4}{*}{0.008} & \multirow{4}{*}{0.042} & \multirow{4}{*}{0.02} \\
        & Male & 71.8 & 69.7 & 70.5 & & & \\
        & Avg.$\uparrow$ & 73.1 & 67.9 & 70.0 & & & \\
        & Diff.$\downarrow$ & 2.6 & 3.7 & 0.9 & & & \\
    \midrule
        \multirow{4}{*}{MFD} & Female & 77.0 & 69.7 & 72.6 & \multirow{4}{*}{0.005} & \multirow{4}{*}{0.051} &\multirow{4}{*} {0.024} \\
        & Male & 77.2 & 72.6 & 74.4 & & & \\
        & Avg.$\uparrow$ & 77.1 & 71.2 & 73.5 & & & \\
        & Diff.$\downarrow$ & 0.2 & 2.9 & 1.8 & & & \\
      \midrule
        \multirow{4}{*}{FairPrune} & Female & 77.6 & 71.1 & 73.4 & \multirow{4}{*}{0.007} & \multirow{4}{*}{\textbf{0.026}} & \multirow{4}{*}{0.014} \\
        & Male & 72.1 & 72.5 & 72.0 & & & \\
        & Avg.$\uparrow$ & 74.8 & 71.8 & 72.7 & & & \\
        & Diff.$\downarrow$ & 5.5 & 1.4 & 1.4 & & & \\
         \midrule
        \multirow{4}{*}{Proposed} & Female & 75.19 & 73.23 & 74.00 & \multirow{4}{*}{\textbf{0.005}} & \multirow{4}{*}{0.030} & \multirow{4}{*}{\textbf{0.01}} \\
        & Male & 78.10 & 71.06 & 73.88 & & & \\
        & Avg.$\uparrow$ & 76.64 & 72.14 & \textbf{73.94} & & & \\
        & Diff.$\downarrow$ & 2.91 & 02.17 & \textbf{0.12} & & & \\
        \bottomrule
    \end{tabular}
\end{table}

\section{Conclusion}
We propose a novel bias mitigation approach that leverages two biased teachers to train a fair student model with improved accuracy. To achieve this, we utilize a multi-weighted loss function that controls model bias based on available sensitive attributes through loss component adjustments to facilitate fair knowledge transfer. Through extensive experimentation, we demonstrate the superiority of the proposed approach in controlling and adjusting the model bias so that the utility of the model does not suffer. Our approach sets a new benchmark for the fairness-accuracy trade-off on two dermatology datasets, i.e., Fitzpatrick-17k and ISIC-2019, and surpasses the existing state-of-the-art methods. In future work, we aim to extend our approach to address bias issues in datasets with multiple sensitive attributes. The proposed approach can also be extended for other use cases with different data modalities, including Magnetic Resonance Imaging (MRI) and computed Tomography (CT) for other medical use cases.

\section{Acknowledgement}
This research received funding from IIT Roorkee and the University Grants Commission (UGC) of India under grant number 190510040512. We also acknowledge the National Supercomputing Mission (NSM) for providing computing resources of ‘PARAM Ganga’ at the Indian Institute of Technology Roorkee, which is implemented by C-DAC and supported by the Ministry of Electronics and Information Technology (MeitY) and Department of Science and Technology (DST), Government of India.

\bibliographystyle{elsarticle-num-names} 



\end{document}